\documentclass[letterpaper, 10 pt, conference]{ieeeconf}
\IEEEoverridecommandlockouts
\makeatletter

\let\proof\@undefined
\let\endproof\@undefined
\makeatother

\usepackage[sort,compress]{cite}

\usepackage{graphicx} 
\graphicspath{{figures/}}%
\usepackage{array}
\usepackage[font=small]{caption}
\usepackage{subcaption}
\usepackage{optidef} 
\usepackage{booktabs}
\usepackage{amsmath} 
\usepackage{color}

\usepackage[ruled]{algorithm2e}
\usepackage{algorithmic}
\usepackage{blindtext}
\usepackage{lipsum}
\usepackage{url}

\usepackage{amsthm}
\theoremstyle{definition}

\usepackage{tikz}
\usetikzlibrary{positioning}

\newcommand{\todo}[1]{{\color{blue} ToDo: #1}}

\definecolor{mypink}{rgb}{0.858, 0.188, 0.478}

\title{\LARGE \bf
On the Impact of Interruptions During Multi-Robot Supervision Tasks}

\author{Abhinav Dahiya, Yifan Cai, Oliver Schneider, Stephen L.\ Smith
\thanks{This work is supported in part by the Natural Sciences and Engineering Research Council of Canada (NSERC)}
\thanks{A. Dahiya, Y. Cai, and S. L. Smith are with Department of Electrical and Computer Engineering and  O. Schneider is with the Department of Management Sciences at the University of Waterloo, Waterloo ON N2L 3G1, Canada. (email: \protect\url{{abhinav.dahiya, yifan.cai, oliver.schneider, stephen.smith}@uwaterloo.ca}).}%
}

\begin{document}

\maketitle

\begin{abstract}
Human supervisors in multi-robot systems are primarily responsible for monitoring robots, but can also be assigned with secondary tasks.  These tasks can act as interruptions and can be categorized as either intrinsic, i.e., being directly related to the monitoring task, or extrinsic, i.e., being unrelated.
In this paper, we investigate the impact of these two types of interruptions through a user study ($N=39$), where participants monitor a number of remote mobile robots 
while intermittently being interrupted by either a robot fault correction task (intrinsic) or a messaging task (extrinsic).
We find that task performance of participants does not change significantly with the interruptions but depends greatly on the number of robots. However, interruptions result in an increase in perceived workload, and extrinsic interruptions have a more negative effect on workload across all NASA-TLX scales.  
Participants also reported switching between extrinsic interruptions and the primary task to be more difficult compared to the intrinsic interruption case.  Statistical significance of these results is confirmed using ANOVA and one-sample t-test.
These findings suggest that when deciding task assignment in such supervision systems, one should limit interruptions from secondary tasks, especially extrinsic ones, in order to limit user workload.
\end{abstract}


\section{Introduction}

Many of the systems built for assisting humans in remote supervision of multiple robots are developed around the assumption that human operators will be solely working on the supervision task \cite{wong2017workload, dahiya2023survey}. However, as robot technology advances, 
the reliance on strict human supervision is reduced. 
This has enabled the development of semi-autonomous robotic systems where the robots can execute most of their tasks autonomously, only requiring human assistance when they encounter some critical states or an unforeseen fault \cite{zanlongo2021scheduling, swamy2020scaled}.  

In a common implementation of such systems, there are several different tasks that a supervisor can be working on (or switching between) at a given time.  Their primary task is to monitor the robots looking for fault status in their operation.  As a secondary task, they may be responsible for resolving faults when robot's automatic correction procedure fails \cite{chien2018attention}. 
Additionally, in a practical scenario, a supervisor may need to work on tasks unrelated to active robot monitoring, such as coordinating with colleagues.  However, even though these secondary tasks are parts of supervisor's job, they can act as interruptions as they take the supervisor's attention away from the primary task of monitoring the robots.  This can be a problem, especially in the case of time-critical systems (e.g., robots navigating on a road network).  
%
%
\begin{figure}[t]
\centering
\vspace{1em}
  \includegraphics[width=0.48\textwidth]{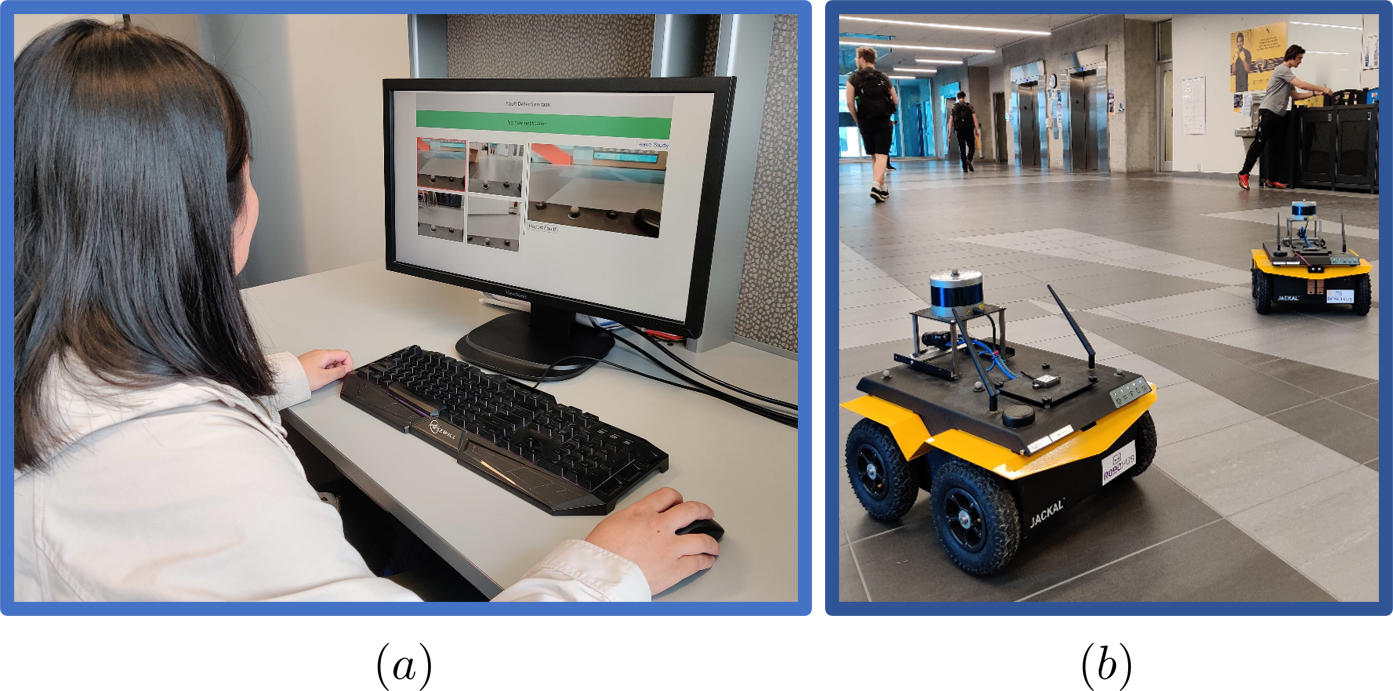}
  \caption{Multi-robot supervision user study setup: a) A participant monitoring multiple robots using the web-based interface, b) Clearpath Jackal robots used for the study.}
  \label{fig:main}
\end{figure}


In the multi-robot supervision literature, it is well-established that controlling a large number of robots negatively affects supervisor's attention and increases their workload \cite{wong2017workload, lewis2013human}.  Researchers have used approaches like implementing different communication strategies \cite{rossi2016supervisory}, using task coordinators \cite{chen2012supervisory} or adjusting robot behaviors \cite{zheng2013supervisory} to tackle this problem.  
%
Several studies from outside the robotics literature have compared the impact of interruptions on workload based on their relation to the primary task \cite{wiegmann2007disruptions, herrick2020impact, fealy2019clinical}.  
However, there is a gap in research in understanding the role of interruptions in multi-robot supervision systems, and the significance of differentiating interruptions based on their relation to the primary task of robot supervision.

In this paper, we investigate the effects of two types of \textit{interruptions} in a multi-robot supervision system.
We consider a system where users primarily work on a monitoring task (reporting faults in robot behaviour) and intermittently face either of the two types of interruptions: 1) Intrinsic: ones related to the primary task, and 2) Extrinsic: ones unrelated to the primary task.  
Given a primary task, we define intrinsic interruptions as the ones that are closely related to the primary task.  In a robot supervision task, intrinsic interruptions can include teleoperating the robot or resolving robot failures. 
We define extrinsic interruptions as those where the supervisor works on something completely unrelated to the primary task environment. This can either be another part of their job or simply an unexpected distraction.

We investigate the effects of the two types of interruptions using a user study with a simulated robot supervision task (Fig.~\ref{fig:main}). The main findings of this work are as follows.  We found the number of robots monitored by the participants to be a good predictor of their performance, both in terms of percentage fault reported ($F=20.23$, $p=0.0$) and average response time ($F=852, p=0.0$).  Even though the interruptions do not significantly affect performance, they do result in an increase in participants' perceived workload on most of the NASA-TLX scales.  Pairwise comparison of different test conditions reveal a significant increase in participants' workload, with extrinsic interruptions resulting in higher workload than intrinsic ones.

The rest of the paper is organized as follows: In Section~\ref{sec:background}, we present some of the existing work on interruptions in workplace, and in multi-robot supervision tasks. In Section~\ref{sec:method}, we describe different elements of our study design and in Section~\ref{sec:results}, the results are presented. The paper ends with Section~\ref{sec:discussion} discussing the findings and implications.

\section{Background and Related Work}
\label{sec:background}

In this section, we discuss existing work on interruptions in workplace and their role in human--multi-robot systems, as well as the relevance of intrinsic and extrinsic interruptions.

\subsection{Human Supervision of Multiple Robots}
Even though robotic systems are rapidly increasing their autonomous capabilities, human supervision is still considered necessary to ensure that task goals are met during unanticipated events \cite{zigoris2003balancing, dahiya2022scalable}.  There exists motivation to decrease the number of supervisors required in large multi-robot systems \cite{squire2010effects}, but doing so negatively impacts supervisors' workload and performance \cite{wong2017workload, chen2010supervisory, riley2005situation}.


The existing research primarily attempts to address this problem from the robotics side of the system, for example, by implementing different communication strategies \cite{rossi2016supervisory}, using task coordinators \cite{chen2012supervisory}, adjusting robot behaviors \cite{zheng2013supervisory} or using frameworks such as sliding autonomy~\cite{music2017control, dias2008sliding}.  However, human factors also play an important role in governing system performance, and it is crucial to design the system in a way that results in effective human operation.     


Interruptions are one of the common issues that can disrupt the ability of operators to maintain attention on a given task.  Studies have shown that interruptions during a task can negatively impact user workload and increase error rates \cite{oury2021building, campoe2017impact}.  However, the role and impact of interruptions in human--multi-robot systems has not been studied adequately in the literature.  This is especially true for remote multi-robot supervision tasks where we naturally have different kinds of interruptions based on their relation to the monitoring task.

\subsection{Interruptions in the Workplace}

Studying interruptions faced by humans is an important area of research in many applications as they can negatively affect the performance and workload of workers \cite{oury2021building, campoe2017impact}.
Interruptions can be simply defined as unanticipated disruption in one's primary task and diversion of attention to a related or unrelated secondary task \cite{sasangohar2012not}. 
Interruptions can originate \textit{externally}, from the environment (noise, notifications or other external factors), or they can arise \textit{internally}, from within the human (due to, for example, boredom or non-task-related thoughts).
They can also be characterized based on their timing, relevance and attentional requirement \cite{addas2015many, gould2014makes, gluck2007matching}.

In the literature on workplace interruptions (mostly in healthcare applications), researchers have studied different types of interruptions based on their relation and relevance to the primary task that workers are performing \cite{wiegmann2007disruptions, herrick2020impact}.  Researchers have used the term \textit{extraneous interruptions} to describe those that do not directly pertain to the primary task, and such interruptions are found to be one of the most common types faced by workers \cite{fealy2019clinical}.  
%


\subsection{Interruptions in Multi-Robot Supervision}
When looking at the multi-robot supervision literature, we notice a gap in the research on how different types of interruptions affect system performance and user workload.  With this paper, we aim to bridge this gap by differentiating different secondary tasks that supervisors need to perform in a multi-robot system into intrinsic and extrinsic interruptions, as ones related and unrelated to the primary task respectively.

This distinction also bears similarities with notion of different types of cognitive loads studied under the Cognitive Load Theory (CLT).  The theory distinguishes between three different types of cognitive load: Intrinsic (load from the task itself), Extrinsic/Extraneous (load not related to the task but induced
by its design), and Germane (load from learner’s deliberate use of cognitive strategies) \cite{young2016measuring, sweller1998cognitive}.  
This distinction provides further motivation to explore significance of differentiating interruptions based on their relationship with the primary task.

{
}
    
\section{Methodology}
\label{sec:method}
In this section, we provide details of our user study, including the application designed for robot monitoring, different tasks that participants encounter, and our study hypotheses.
The study adopts a mixed factorial design in which the type of interruption (intrinsic or extrinsic) is the \textit{within-participant} factor, while the number of robots to be monitored ($4$ or $9$) is the \textit{between-participant} factor.  These numbers are selected based on a pilot study, which ensures that the difficulty level of the monitoring task ranges from easy to moderately difficult, and aligns with the existing understanding of human psychological attention limit. Moreover, similar numbers have been used in prior studies on human-multi-robot systems across various applications \cite{chen2014human}.  We recruited $39$ participants in total distributed evenly between $4$ and $9$ robots cases.  The participants of the study consisted of university students and individuals who were recruited via personal networks. None of the participants had prior experience using a robot monitoring interface.  The study has been reviewed and received ethics clearance through a University of Waterloo Research Ethics Committee (ORE\#43628).
\begin{figure*}[ht]
\centering
  \frame{\includegraphics[width=0.7\textwidth]{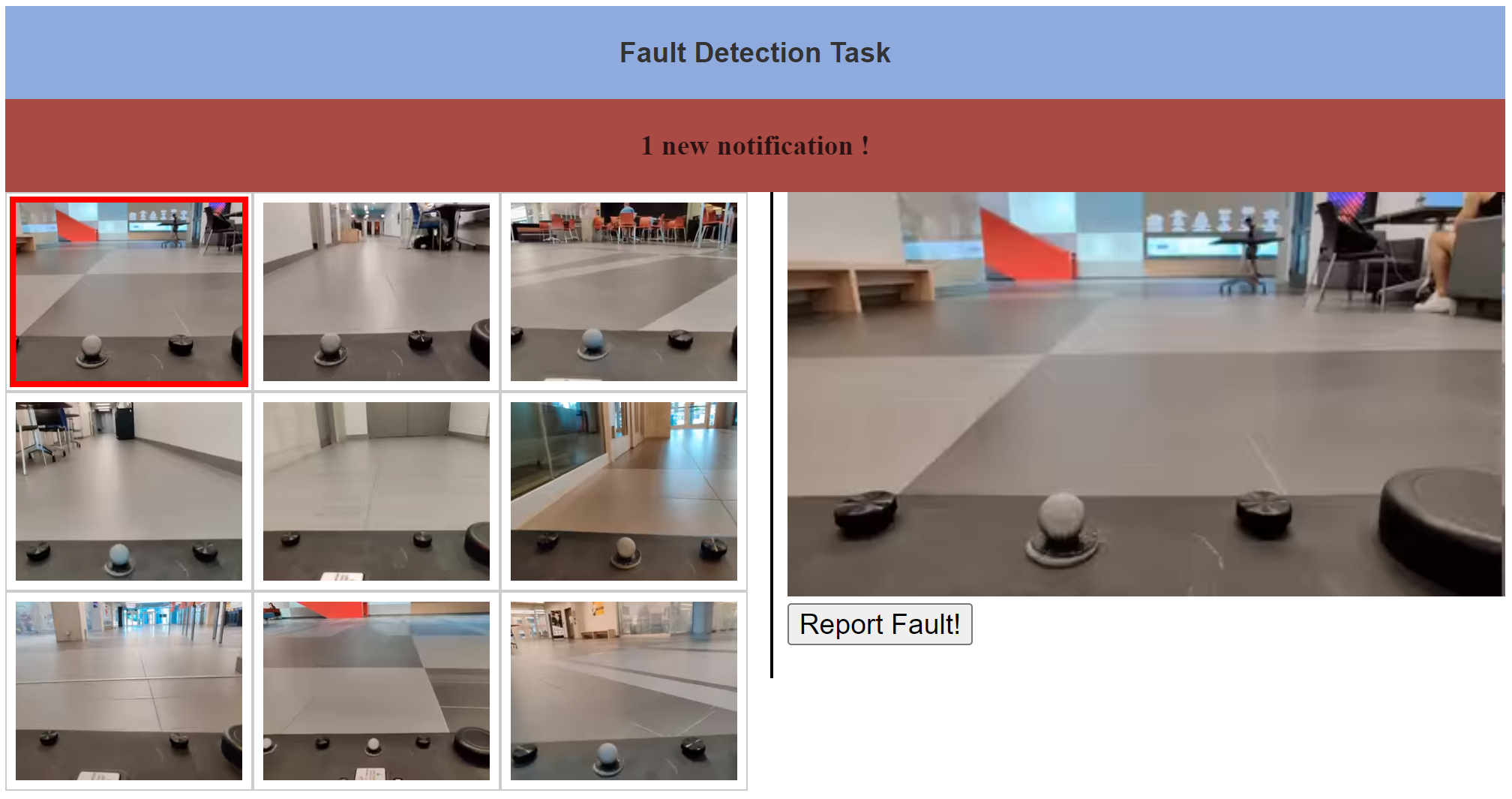}}
  \caption{Interface for monitoring. The grid of robot cards on the left side shows video information from all robots. On the right, there is an enlarged visual of the selected robot. On the top, there is a notification panel for incoming secondary tasks.}
  \label{fig:monitor}
\end{figure*}
\subsection{Study Design}
\label{sec:study}
A web-based application is designed to conduct the study.  The application replicates a basic setup of a remote-monitoring interface, with camera feed from multiple robots in one half of the screen, and an enlarged view of a single robot in the other half.  Participants can select any one robot for an enlarged view for detailed inspection and for reporting faults in that robot.  The application also displays notifications for any interruption that may arrive, which in our case are the prompts for secondary tasks based on the test condition.  

Each robot's \textit{camera feed} shows a pre-recorded video from a camera mounted on the robot, navigating in an indoor building environment with light foot traffic (see Fig.~\ref{fig:main}(b)).  The robot navigation was intentionally corrupted to include faults, which were designed to appear as one of three behaviors in robots' movements: 1) Stops moving, 2) Moving in circles at a spot, and 3) Turning side-to-side without moving forward.
The faults were randomly introduced during robot navigation with their start and end times determined randomly as well.  Each fault lasted for at least $20$ seconds and there was at least $30$ seconds between two faults. In the videos used for the study, the robots experienced faults between $1$ and $5$ times, with an average fault duration of $30$ seconds. A robot was in a fault state for about $29.5\%$ of the total duration."


At any given time during the experiment, a participant can be working on one of the following three tasks:

\textbf{1) Robot monitoring}:  This is the primary (default) task during the experiment, which requires participants to monitor all the robots shown in the interface (Fig.~\ref{fig:monitor}), and detect if any of the robots is in fault state.  Once a fault is detected, participants need to select that robot and press the `Report Fault' button.  This action is programmed to \textit{fix} the fault, and its camera feed is refreshed to show the robot navigating normally again.  Participants are required to report all the faults that appear during the experiment and as soon as possible from their appearance. 
%

\begin{figure}[ht]
\centering
  \frame{\includegraphics[width=\columnwidth]{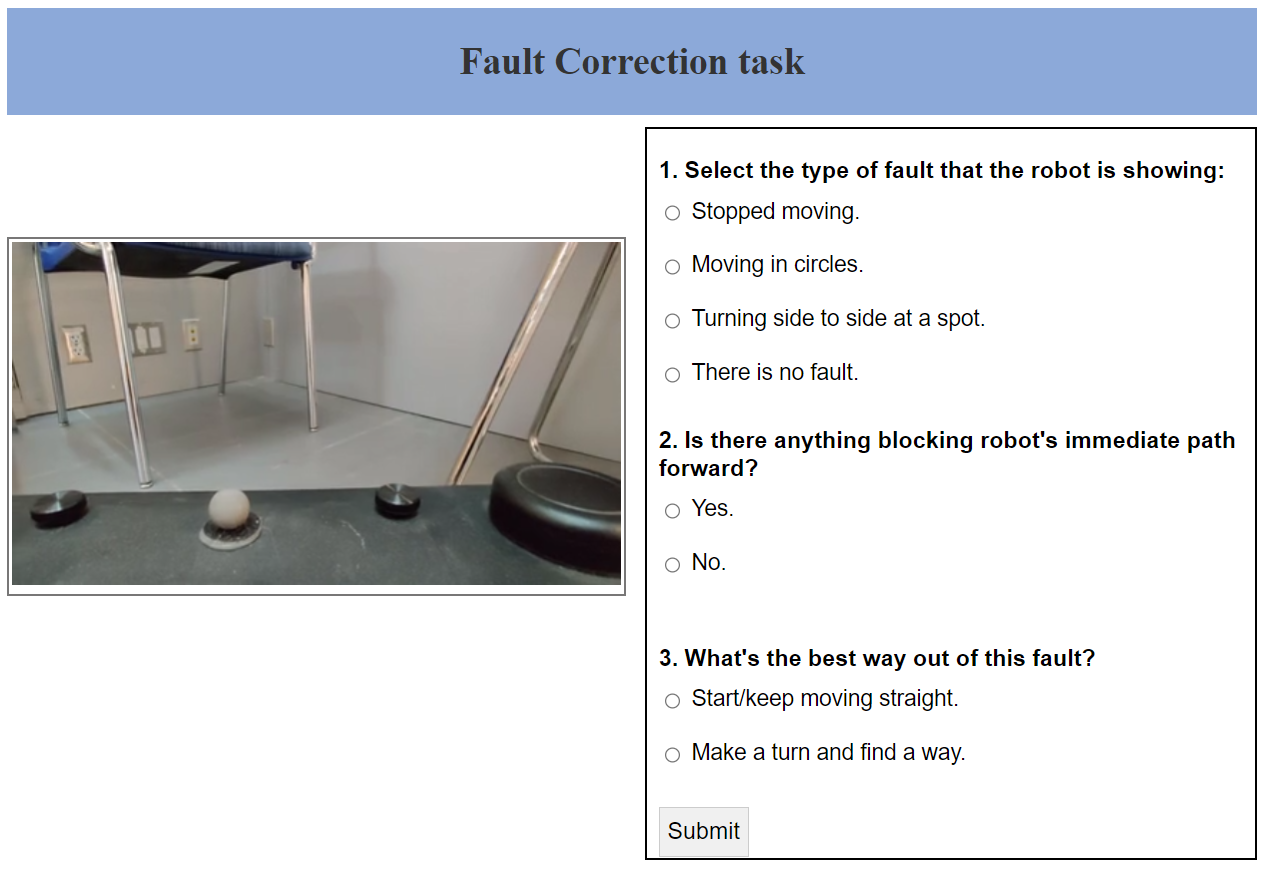}}
  \caption{Interface for fault correction task. On the left, participants see a robot potentially in a fault state. On the right, there are several questions to characterize the fault.}
  \label{fig:correction}
\end{figure}
\textbf{2) Fault Correction}:  
During this secondary task, participants are shown a video feed of a potentially faulty robot and are required to answer questions about it (see Fig.~\ref{fig:correction}). This acts as an intrinsic interruption closely related to the primary monitoring task while not requiring any technical knowledge about the robot operation.
Once all questions are answered correctly, the participant is taken back to the monitoring task.  For the study, videos for this task are randomly selected from a pool of $15$ videos each showing a different type of fault (or no fault at all).

\begin{figure}[ht]
\centering
  \frame{\includegraphics[width=\columnwidth]{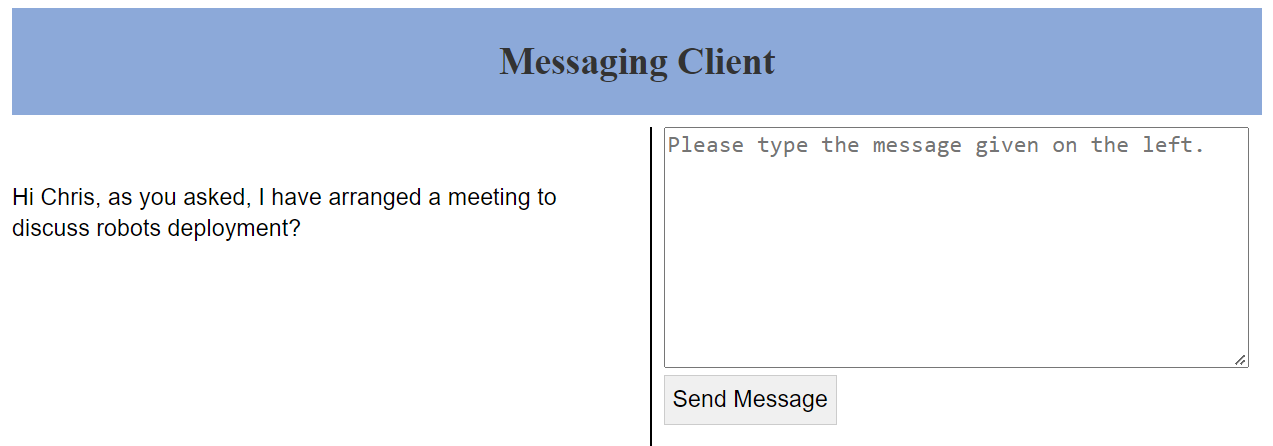}}
  \caption{Interface for the messaging task. On the left, participants see a pre-written message to be sent. On the right is a text box to type the message and the send button.}
  \label{fig:message}
\end{figure}

\textbf{3) Messaging Task}:  This secondary task represents the extrinsic interruption during which participants are required to write a message to their colleagues.  The message is already displayed on the screen and participants need to type it again in the space provided (see Fig.~\ref{fig:message}).  Once the message is typed, the participant can press the `Send message' button and is then taken back to the monitoring task.  For the study, messages for this task are randomly selected from a pool of $15$ messages, each with $85$ or fewer characters.

\subsection{Procedure}
Each participant first goes through a training session and then completes the experiment under three different test conditions, which decide the type of interruption they will be facing.  The order of these three conditions is counterbalanced.  Therefore participants see these conditions in a different order, with a minute-long break between conditions.  The three conditions are described below.

\textbf{1) Condition-0 (No interruption)}: In this condition, no interruption occurs and participants work on the robot monitoring task for the whole duration.
    
\textbf{2) Condition-1 (Intrinsic interruption)}: In this condition, participants are shown a notification after they spend certain amount of time on the monitoring task.  In this condition, clicking a notification takes the participant to the fault correction task screen.
    
\textbf{3) Condition-2 (Extrinsic interruption)}:  Under this condition, the experiment is conducted in a similar way as Condition-1, except that clicking notifications takes the participant to the messaging task screen.

Under each condition, participants work on the monitoring task for $2$ minutes, during which they may receive notifications for interruptions in the form of secondary tasks.  These notifications are randomly presented after the participant spends between $15$ and $40$ seconds on the monitoring task, with the exact time sampled from a uniform distribution.  These durations were selected based on pilot testing to ensure a balance between time spent on the primary and secondary tasks.

\subsection{Metrics}
After a participant is finished with the assigned tasks, they are asked to fill out the NASA-TLX questionnaire \cite{hart2006nasa} (once after each test condition). Once they complete the task under all conditions, they fill a post-experiment questionnaire and the procedure is finished.
Additionally, the application also records participant's performance parameters, such as faults reported and response time of identifying faults.  {If a robot gets into fault when a participant is working on an interruption, the response time only starts to count when they resume the monitoring task.  This allows us to avoid counting the time a participant spends on an interruption task towards their response time.}

    
 
 
 \subsection{Hypotheses}
This study seeks to learn how the intrinsic and extrinsic interruptions can affect supervisor's performance and workload in a multi-robot remote supervision system.  
However, NASA-TLX scores are highly influenced by individual differences especially when using the unweighted scores\footnote{We do not use category weights for two reasons:  First, it is unclear how using category weights affect sensitivity of the score across different systems \cite{hart2006nasa}.  Second, administering ranking questions for the weighting step requires a fair amount of effort from the participants which, in our case, is comparable to the effort required for the task itself.} \cite{hart2006nasa}.  To eliminate the effects of individual differences, we analyze change in scores of participants across test conditions.

For this study, we propose the following null hypotheses:\\
\textbf{$\boldsymbol{H_0}$-1:}
Task performance does not differ across test conditions (type of interruptions shown).\\
\textbf{$\boldsymbol{H_0}$-2:}
Perceived workload does not differ across test conditions.\\


    
        

\section{Results}
\label{sec:results}
We analyze the experiment data under three categories: users' performance, perceived workload, and responses to post-experiment questionnaire.




\subsection{Results on Performance}
For the presented task, we use the following metrics as a measure of performance:  First is the percentage of faults reported, calculated as the ratio of faults reported to total faults appeared during a task.  Second is the response time, calculated as the average time it took a user to report a fault (time from appearance of a fault to its reporting\footnote{If a participant fails to report a fault, the fault duration is considered to be the response time for that fault.}).  

\begin{figure}[ht]
      \centering
      \includegraphics[width=\columnwidth]{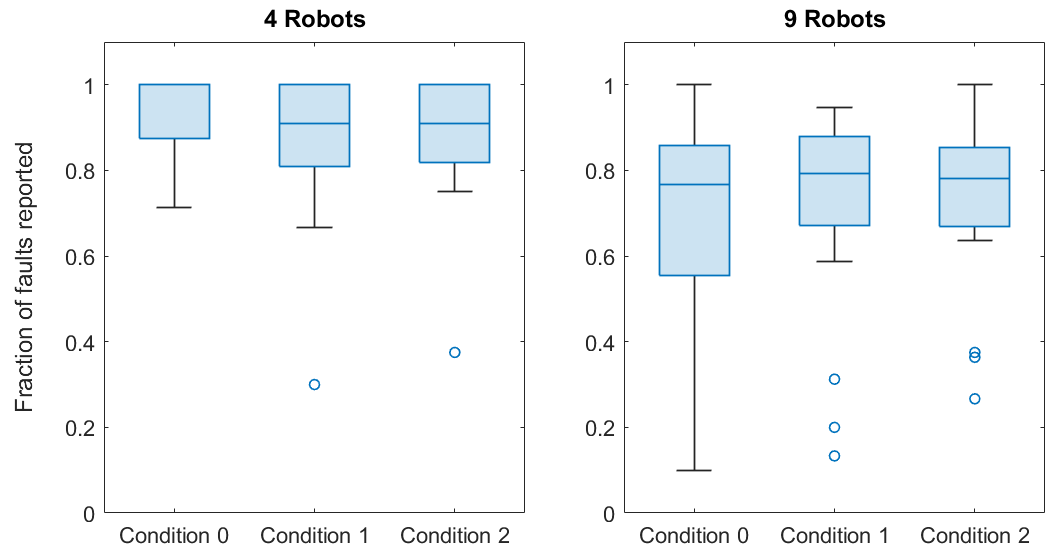}
      \caption {Percentage Fault Reported under different test conditions for participants monitoring $4$ robots (left) and $9$ robots (right).}
      \label{fig:performance_percentFault}
   \end{figure}

Figure~\ref{fig:performance_percentFault} shows the percentage of faults reported by participants under each test condition.  From the graph, we observe that when monitoring $4$ robots, most of the participants were able to detect majority of the faults ($>70\%$) under all three test conditions.  As the number of robots increases to $9$, the percentage of faults reported decreases under all three test conditions.  A 2-Way ANOVA confirms that number of robots is a very strong predictor of percent fault reported ($F=20.23, p=0.0$).  This is expected as participants are required to keep attention over a larger stream of information.  The conditions themselves do not show any conclusive effect on the outcome ($F=0.21, p=0.81$).   

\begin{figure}[ht]
      \centering
      \includegraphics[width=\columnwidth]{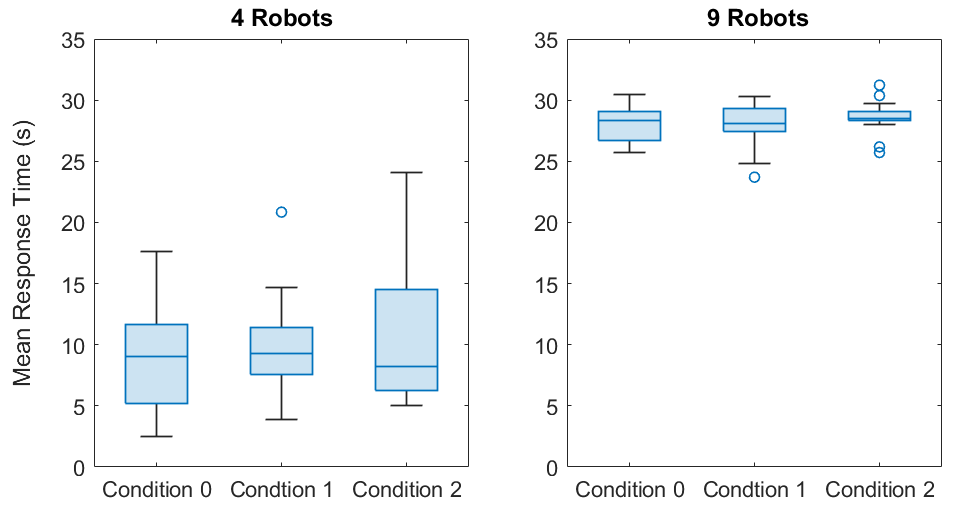}
      \caption {Average response time (time from fault appearance to fault reporting) for all users under different conditions.}
      \label{fig:performance_responseTime}
  \end{figure}

Figure~\ref{fig:performance_responseTime} shows the average amount of time a participant took to report faults under different conditions.  These results are similar to the case of percent fault reported, where number of robots are a strong predictor of outcome ($F=852, p=0.0$) while interruption type does not have a significant effect ($F=0.05, p=0.95$).  
{This may be partially because the response time only starts to count when participants resume the monitoring task.  This allows us to avoid counting the time a participant spends on an interruption task towards their response time.}

Given these results, \textbf{the null hypothesis $\boldsymbol{H_0}$-1 cannot be rejected}: task performance does not differ with test conditions (type of interruptions shown).

{
 }

\subsection{Results on Workload}
\begin{figure}[ht]
      \centering
      \includegraphics[width=\columnwidth]{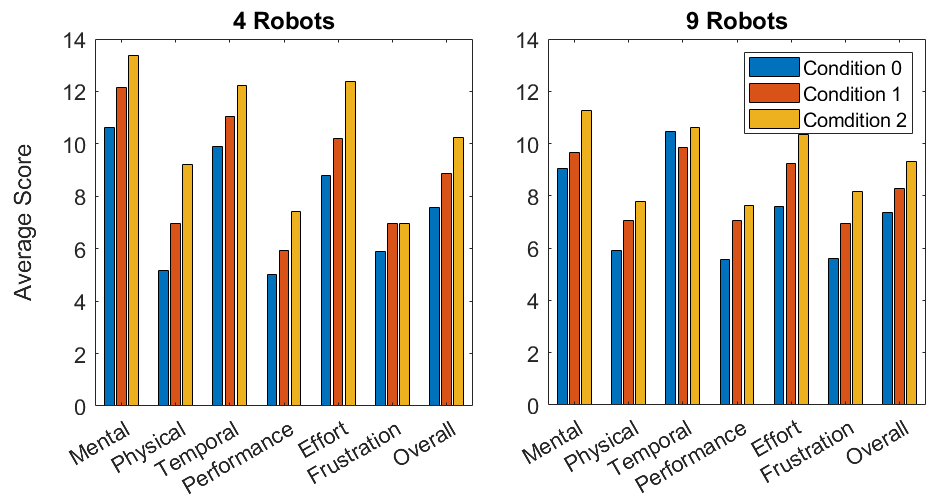}
      \caption {Average Rating for TLX Questions for different conditions for $4$ and $9$ robots.}
      \label{fig:workload_TLX}
  \end{figure}

Figure~\ref{fig:workload_TLX} shows the participants' perceived workload during different test conditions, measured as NASA-TLX scores.
   
From the figure, we observe that participants reported higher workload, on average, under Condition-2 (extrinsic interruptions) on most of the workload categories followed by Condition-1 and then Condition-0, regardless of the number of robots they monitored.  

Since we are using unweighted scores, it is more relevant for the study to compare change in scores between test conditions for individual participants rather than taking average.  Therefore, 
we present pairwise comparison of participants' scores for each workload category.  

      \begin{figure}[ht]
      \centering
      \includegraphics[width=0.9\columnwidth]{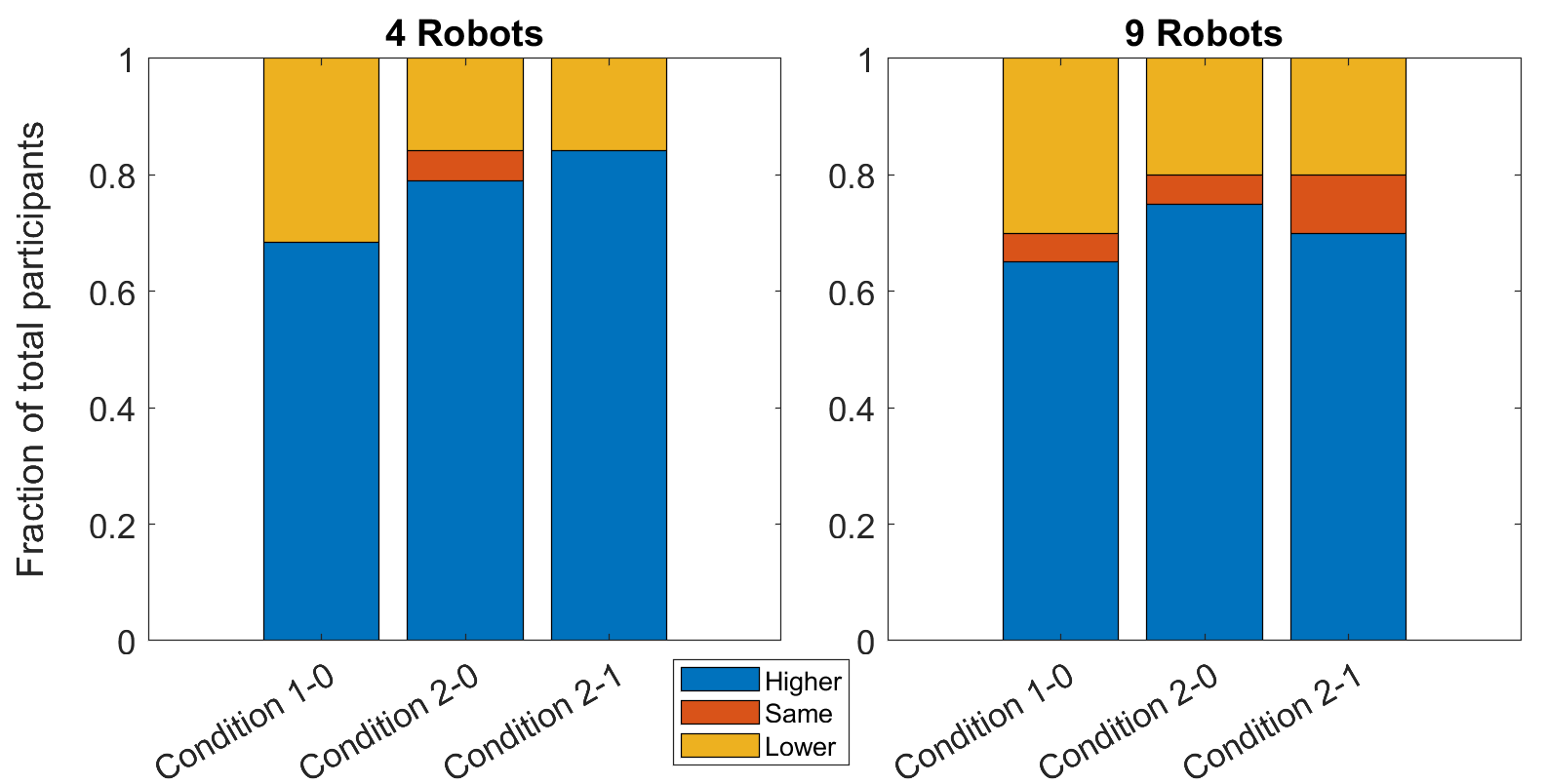}
      \caption {Fraction of participants who reported higher/lower/same scores between two conditions.  For example, for pair 1-0, blue represents that participants' reported higher workload in condition-1, orange means scores were same for both conditions and yellow means a higher workload reported in condition-0.}
      \label{fig:stackedBarOverallWorkload}
  \end{figure}

Figure~\ref{fig:stackedBarOverallWorkload} shows how individual participant's workload scores changed between conditions.  These are shown as the percentage of participants who reported higher, lower, or the same workload score.  Comparisons are performed in going from Condition-0 to 1, Condition-0 to 2, and Condition-1 to 2.
A one sample t-test reveals that differences between most of the test conditions are significantly different from a zero-mean distribution, except for the difference between condition-1 and condition-0 for the 9-robot case. 
Table~\ref{table:t-test_workload} shows the result of these t-tests.  

\begin{table}[h]
\caption{One-sample t-test p-values for different pairwise comparison of TLX scores.}
\label{table:t-test_workload}
\renewcommand{\arraystretch}{1.3}
\begin{center}
\begin{tabular}{@{}lccc@{}}
\toprule
& Conditions 1-0 & Conditions 2-0 & Conditions 2-1 \\ 
\midrule
 $4$ robots & $0.0299$ & $0.0009$ & $0.0011$ \\
 $9$ robots & $0.1150$ & $0.0109$ & $0.0419$ \\
 Overall & $0.0071$ & $0.0000$ & $0.0001$\\
\bottomrule
\end{tabular}
\end{center}
\end{table}



\begin{figure}[ht]
  \centering
  \includegraphics[width=0.84\columnwidth]{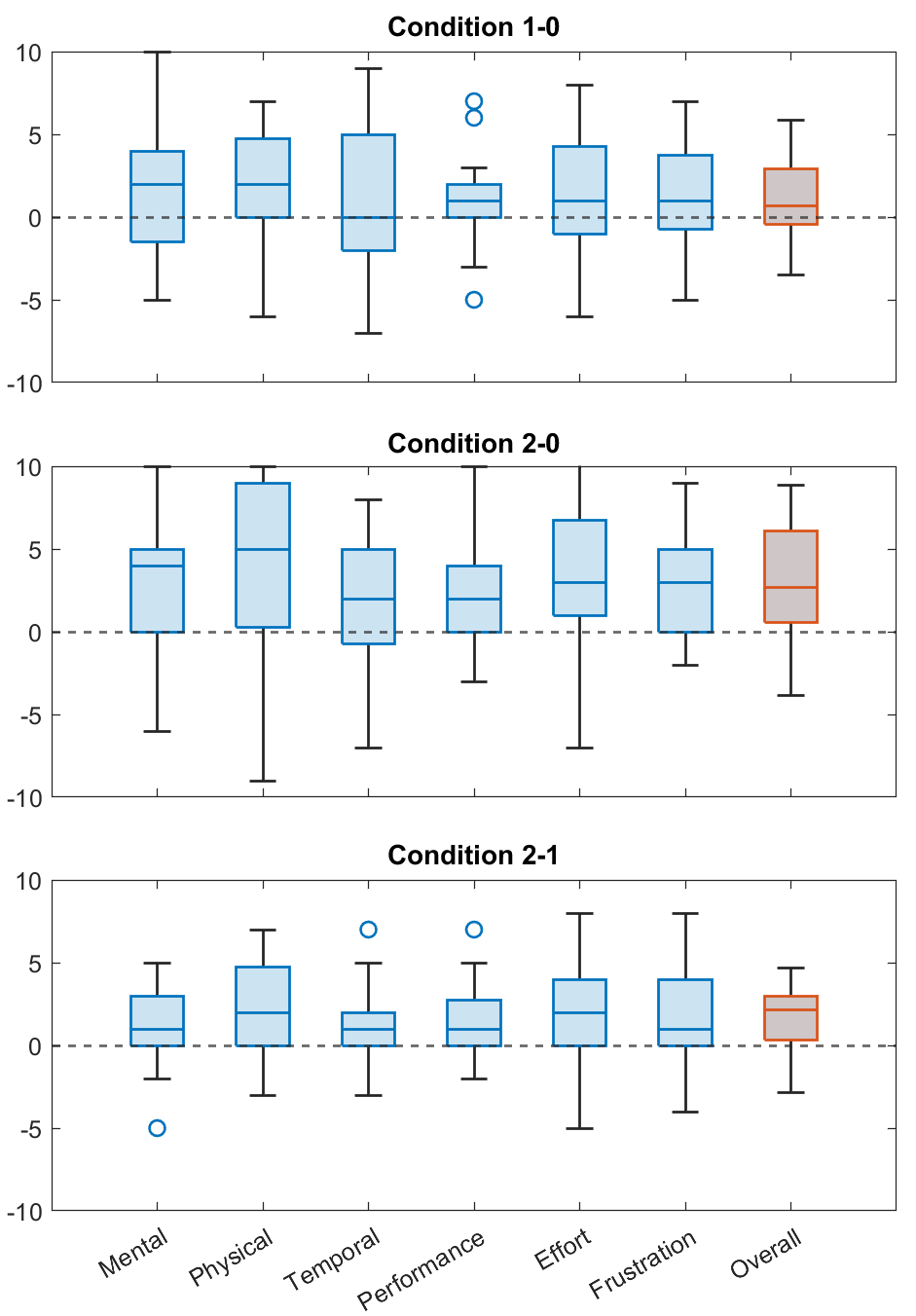}
  \caption {Difference in TLX scores between different conditions for $4$ robots.  For example, the Condition 1-0 plots show distribution of condition 1 minus condition 0 scores for individual participant.}
  \label{fig:workload_TLX_box4}
\end{figure}
\begin{figure}[ht]
  \centering
  \includegraphics[width=0.85\columnwidth]{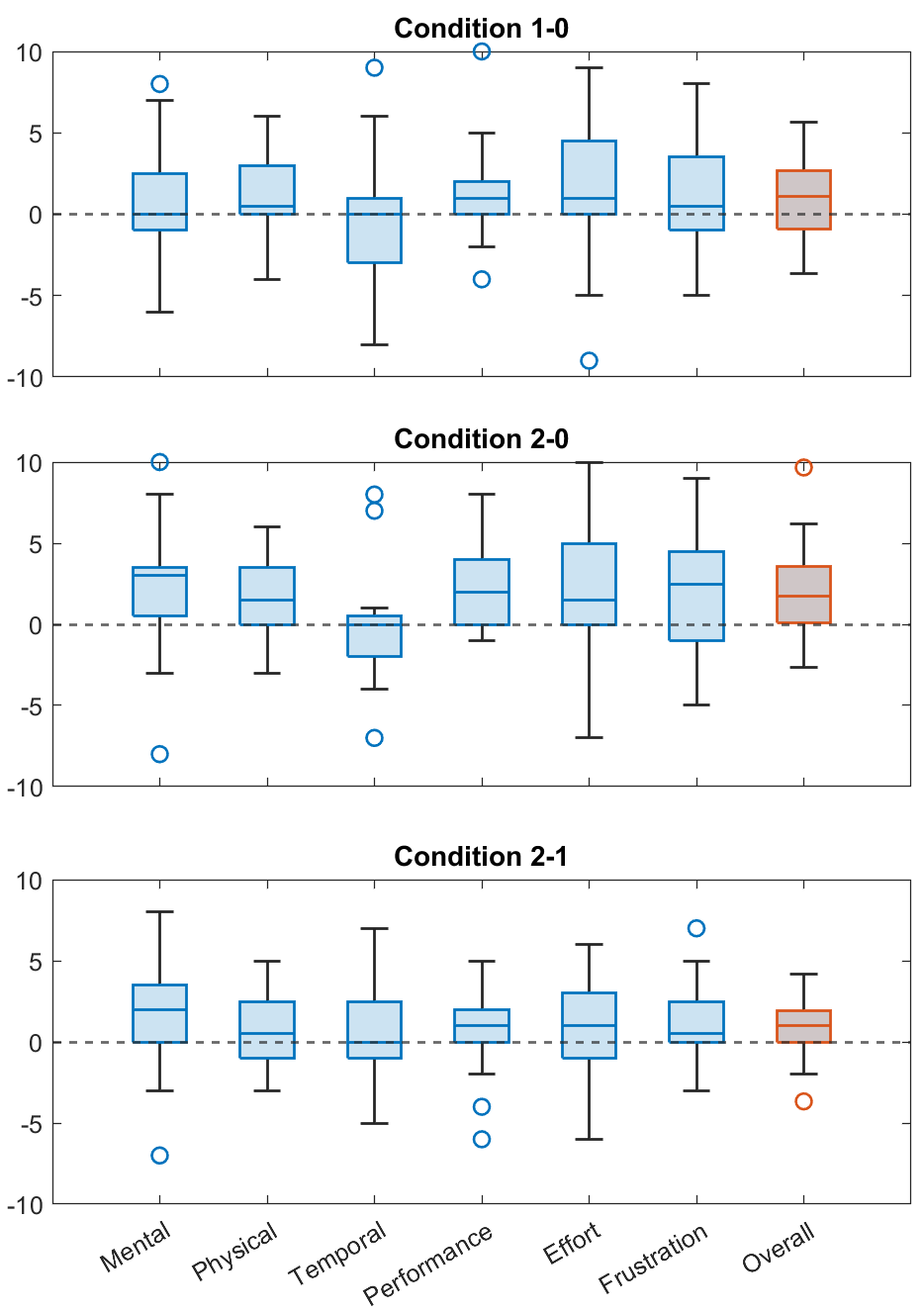}
  \caption {Difference in TLX scores between different conditions for $9$ robots.}
  \label{fig:workload_TLX_box9}
\end{figure}

To provide a more comprehensive analysis, we also present the distribution of changes in workload scores for participants across all workload categories and pairwise comparisons in Figures~\ref{fig:workload_TLX_box4} and~\ref{fig:workload_TLX_box9}. The results indicate that a majority of participants reported higher workload scores under condition-2 compared to both condition-0 and condition-1.

These findings lead us to \textbf{reject the null hypothesis $\boldsymbol{H_0}$-2}, i.e., perceived workload differs with test conditions (type of interruptions).

\subsection{Post-Experiment Questionnaire}
Besides measuring performance and workload of the participants, we also asked them some further questions regarding their perception of different tasks they performed during the whole experiment.  This post-experiment questionnaire is shown in Table~\ref{table:postExp}, and participants' agreement with each statement was recorded on a 20-point scale (1 being Strongly Disagree and 20 being Strongly Agree).  
\begin{table}[h!]
\setlength{\arrayrulewidth}{0.2mm}
\setlength{\tabcolsep}{5pt}
\renewcommand{\arraystretch}{1.2}
\centering
\caption{Prompts used in post-experiment questionnaire.}
\label{table:postExp}
\begin{tabular}{ | p{0.25cm} | p{7cm}| } 
 \hline
 1 & I found the fault correction tasks and messaging tasks disruptive while monitoring the robots. \\ 
 2 & I found it difficult to switch from the robot monitoring task to the correction task. \\
 3 & I found it difficult to switch from the robot monitoring task to the messaging task. \\
 4 & I found it difficult to resume the robot monitoring task after the correction task. \\
 5 & I found it difficult to resume the robot monitoring task after the messaging task. \\
 \hline
\end{tabular}
\end{table}
 
From participants' responses, we observe that majority of participants found the interruption tasks disruptive while monitoring the robots, with higher average score among participants who monitored $9$ robots (Question 1).  Participants also reported more difficulty while switching from monitoring task to the messaging task compared to the fault correction task, with mean difference $\bar{\mu} = 2.79$ (Questions 2,3).  A one sample t-test confirms the significance of difference ($p=0.015$).  We note similar results on perceived difficulty for resuming the monitoring task after an interruption (Questions 4,5) ($\bar{\mu}=3.20$, $p=0.001$).



\section{Conclusion and Discussion}
\label{sec:discussion}
The study reveals some interesting features of multi-robot supervision systems where users monitor a number of independent mobile robots.
From the results, we observe that while working on the robot monitoring task, any interruption, be it intrinsic (fault correction) or extrinsic (messaging), will result in an increase of user workload.  The effects of extrinsic interruptions on workload are more severe than those of intrinsic ones for both four and nine robots cases.  

However, the impact of these interruptions on the task performance is found to be insignificant.  The number of robots being monitored is observed to be the major factor in a change of performance.  One possible reason for this observation is that the monitoring task in our system requires a short working memory as faults in the system are determined solely based on the current state/short-term behaviour of the robots.  
%
Even though the type of interruptions does not significantly affect performance, participants reported them to be disruptive while monitoring the robots.  Participants also reported that switching to the extrinsic task and resuming the monitoring task afterwards is more difficult than for the intrinsic task.

These findings suggest that when designing such multi-robot supervision systems, it is important to prevent interruptions from extrinsic tasks while working on robot monitoring.  It may be helpful to postpone such interruptions towards the end of the task. It may also be helpful to distribute the responsibility of robot monitoring and fault correction tasks among different operators to limit supervisors switching between the two tasks. 
%
%
%
In future, we would like to expand this study to further characterize the effects of intrinsic and extrinsic interruptions by controlling interruption frequency, changing task difficulty, and having a larger number of robots to monitor.  It is also interesting to explore how the results will change if the secondary tasks are introduced in a multi-tasking scenario instead of being separate interruption tasks, where users try to work on different tasks simultaneously.


\begin{thebibliography}{10}
\providecommand{\url}[1]{#1}
\csname url@rmstyle\endcsname
\providecommand{\newblock}{\relax}
\providecommand{\bibinfo}[2]{#2}
\providecommand\BIBentrySTDinterwordspacing{\spaceskip=0pt\relax}
\providecommand\BIBentryALTinterwordstretchfactor{4}
\providecommand\BIBentryALTinterwordspacing{\spaceskip=\fontdimen2\font plus
\BIBentryALTinterwordstretchfactor\fontdimen3\font minus
  \fontdimen4\font\relax}
\providecommand\BIBforeignlanguage[2]{{%
\expandafter\ifx\csname l@#1\endcsname\relax
\typeout{** WARNING: IEEEtran.bst: No hyphenation pattern has been}%
\typeout{** loaded for the language `#1'. Using the pattern for}%
\typeout{** the default language instead.}%
\else
\language=\csname l@#1\endcsname
\fi
#2}}

\bibitem{wong2017workload}
C.~Y. Wong and G.~Seet, ``Workload, awareness and automation in multiple-robot
  supervision,'' \emph{International Journal of Advanced Robotic Systems},
  vol.~14, no.~3, p. 1729881417710463, 2017.

\bibitem{dahiya2023survey}
A.~Dahiya, A.~M. Aroyo, K.~Dautenhahn, and S.~L. Smith, ``A survey of
  multi-agent human--robot interaction systems,'' \emph{Robotics and Autonomous
  Systems}, vol. 161, p. 104335, 2023.

\bibitem{zanlongo2021scheduling}
S.~A. Zanlongo, P.~Dirksmeier, P.~Long, T.~Padir, and L.~Bobadilla,
  ``Scheduling and path-planning for operator oversight of multiple robots,''
  \emph{Robotics}, vol.~10, no.~2, p.~57, 2021.

\bibitem{swamy2020scaled}
G.~Swamy, S.~Reddy, S.~Levine, and A.~D. Dragan, ``Scaled autonomy: Enabling
  human operators to control robot fleets,'' in \emph{2020 IEEE International
  Conference on Robotics and Automation (ICRA)}.\hskip 1em plus 0.5em minus
  0.4em\relax IEEE, 2020, pp. 5942--5948.

\bibitem{chien2018attention}
S.~Y. Chien, Y.~L. Lin, P.~J. Lee, S.~Han, M.~Lewis, and K.~Sycara, ``Attention
  allocation for human multi-robot control: Cognitive analysis based on
  behavior data and hidden states,'' \emph{International Journal of
  Human-Computer Studies}, vol. 117, pp. 30--44, 2018.

\bibitem{lewis2013human}
M.~Lewis, ``Human interaction with multiple remote robots,'' \emph{Reviews of
  Human Factors and Ergonomics}, vol.~9, no.~1, pp. 131--174, 2013.

\bibitem{rossi2016supervisory}
A.~Rossi, M.~Staffa, and S.~Rossi, ``Supervisory control of multiple robots
  through group communication,'' \emph{IEEE Transactions on Cognitive and
  Developmental Systems}, vol.~9, no.~1, pp. 56--67, 2016.

\bibitem{chen2012supervisory}
J.~Y. Chen and M.~J. Barnes, ``Supervisory control of multiple robots: Effects
  of imperfect automation and individual differences,'' \emph{Human Factors},
  vol.~54, no.~2, pp. 157--174, 2012.

\bibitem{zheng2013supervisory}
K.~Zheng, D.~F. Glas, T.~Kanda, H.~Ishiguro, and N.~Hagita, ``Supervisory
  control of multiple social robots for navigation,'' in \emph{2013 8th
  ACM/IEEE International Conference on Human-Robot Interaction (HRI)}.\hskip
  1em plus 0.5em minus 0.4em\relax IEEE, 2013, pp. 17--24.

\bibitem{wiegmann2007disruptions}
D.~A. Wiegmann, A.~W. ElBardissi, J.~A. Dearani, R.~C. Daly, and T.~M.
  Sundt~III, ``Disruptions in surgical flow and their relationship to surgical
  errors: an exploratory investigation,'' \emph{Surgery}, vol. 142, no.~5, pp.
  658--665, 2007.

\bibitem{herrick2020impact}
H.~M. Herrick, S.~Lorch, J.~Y. Hsu, K.~Catchpole, and E.~E. Foglia, ``Impact of
  flow disruptions in the delivery room,'' \emph{Resuscitation}, vol. 150, pp.
  29--35, 2020.

\bibitem{fealy2019clinical}
G.~Fealy, S.~Donnelly, G.~Doyle, M.~Brenner, M.~Hughes, E.~Mylotte,
  E.~Nicholson, and M.~Zaki, ``Clinical handover practices among healthcare
  practitioners in acute care services: A qualitative study,'' \emph{Journal of
  clinical nursing}, vol.~28, no. 1-2, pp. 80--88, 2019.

\bibitem{zigoris2003balancing}
P.~Zigoris, J.~Siu, O.~Wang, and A.~T. Hayes, ``Balancing automated behavior
  and human control in multi-agent systems: a case study in roboflag,'' in
  \emph{Proceedings of the 2003 American Control Conference, 2003.},
  vol.~1.\hskip 1em plus 0.5em minus 0.4em\relax IEEE, 2003, pp. 667--671.

\bibitem{dahiya2022scalable}
A.~Dahiya, N.~Akbarzadeh, A.~Mahajan, and S.~Smith, ``Scalable operator
  allocation for multi-robot assistance: A restless bandit approach,''
  \emph{IEEE Transactions on Control of Network Systems}, pp. 1397--1408, 2022.

\bibitem{squire2010effects}
P.~Squire and R.~Parasuraman, ``Effects of automation and task load on task
  switching during human supervision of multiple semi-autonomous robots in a
  dynamic environment,'' \emph{Ergonomics}, vol.~53, no.~8, pp. 951--961, 2010.

\bibitem{chen2010supervisory}
J.~Y. Chen, M.~J. Barnes, and M.~Harper-Sciarini, ``Supervisory control of
  multiple robots: Human-performance issues and user-interface design,''
  \emph{IEEE Transactions on Systems, Man, and Cybernetics, Part C
  (Applications and Reviews)}, vol.~41, no.~4, pp. 435--454, 2010.

\bibitem{riley2005situation}
J.~M. Riley and M.~R. Endsley, ``Situation awareness in hri with collaborating
  remotely piloted vehicles,'' in \emph{proceedings of the Human Factors and
  Ergonomics Society Annual Meeting}, vol.~49, no.~3.\hskip 1em plus 0.5em
  minus 0.4em\relax SAGE Publications Sage CA: Los Angeles, CA, 2005, pp.
  407--411.

\bibitem{music2017control}
S.~Musi{\'c} and S.~Hirche, ``Control sharing in human-robot team
  interaction,'' \emph{Annual Reviews in Control}, vol.~44, pp. 342--354, 2017.

\bibitem{dias2008sliding}
M.~B. Dias, B.~Kannan, B.~Browning, E.~Jones, B.~Argall, M.~F. Dias, M.~Zinck,
  M.~Veloso, and A.~Stentz, ``Sliding autonomy for peer-to-peer human-robot
  teams,'' in \emph{International conference on intelligent autonomous
  systems}, 2008, pp. 332--341.

\bibitem{oury2021building}
J.~D. Oury and F.~E. Ritter, \emph{Building Better Interfaces for Remote
  Autonomous Systems: An Introduction for Systems Engineers}.\hskip 1em plus
  0.5em minus 0.4em\relax Springer Nature, 2021.

\bibitem{campoe2017impact}
K.~R. Campoe and K.~K. Giuliano, ``Impact of frequent interruption on nurses’
  patient-controlled analgesia programming performance,'' \emph{Human factors},
  vol.~59, no.~8, pp. 1204--1213, 2017.

\bibitem{sasangohar2012not}
F.~Sasangohar, B.~Donmez, P.~Trbovich, and A.~C. Easty, ``Not all interruptions
  are created equal: positive interruptions in healthcare,'' in
  \emph{Proceedings of the Human Factors and Ergonomics Society Annual
  Meeting}, vol.~56, no.~1.\hskip 1em plus 0.5em minus 0.4em\relax SAGE
  Publications Sage CA: Los Angeles, CA, 2012, pp. 824--828.

\bibitem{addas2015many}
S.~Addas and A.~Pinsonneault, ``The many faces of information technology
  interruptions: a taxonomy and preliminary investigation of their performance
  effects,'' \emph{Information Systems Journal}, vol.~25, no.~3, pp. 231--273,
  2015.

\bibitem{gould2014makes}
A.~J. Gould, ``What makes an interruption disruptive? understanding the effects
  of interruption relevance and timing on performance,'' Ph.D. dissertation,
  UCL (University College London), 2014.

\bibitem{gluck2007matching}
J.~Gluck, A.~Bunt, and J.~McGrenere, ``Matching attentional draw with utility
  in interruption,'' in \emph{Proceedings of the SIGCHI Conference on Human
  Factors in Computing Systems}, 2007, pp. 41--50.

\bibitem{young2016measuring}
J.~Q. Young, D.~M. Irby, M.-L. Barilla-LaBarca, O.~Ten~Cate, and P.~S.
  O’Sullivan, ``Measuring cognitive load: mixed results from a handover
  simulation for medical students,'' \emph{Perspectives on medical education},
  vol.~5, no.~1, pp. 24--32, 2016.

\bibitem{sweller1998cognitive}
J.~Sweller, J.~J. Van~Merrienboer, and F.~G. Paas, ``Cognitive architecture and
  instructional design,'' \emph{Educational psychology review}, vol.~10, no.~3,
  pp. 251--296, 1998.

\bibitem{chen2014human}
J.~Y. Chen and M.~J. Barnes, ``Human--agent teaming for multirobot control: A
  review of human factors issues,'' \emph{IEEE Transactions on Human-Machine
  Systems}, vol.~44, no.~1, pp. 13--29, 2014.

\bibitem{hart2006nasa}
S.~G. Hart, ``{NASA}-task load index ({NASA}-tlx); 20 years later,'' in
  \emph{Proceedings of the human factors and ergonomics society annual
  meeting}, vol.~50, no.~9, 2006, pp. 904--908.

\end{thebibliography}
\end{document}